\documentclass[11pt]{article}

\usepackage[letterpaper, margin=1in]{geometry}

\usepackage[utf8]{inputenc}
\usepackage[T1]{fontenc}
\IfFileExists{lmodern.sty}{\usepackage{lmodern}}{\usepackage{type1cm}}
\usepackage[final,expansion=false]{microtype}

\usepackage{amsmath, amssymb}

\usepackage{booktabs}
\usepackage{array}
\usepackage{calc}
\usepackage{longtable}

\usepackage{graphicx}

\usepackage[hidelinks]{hyperref}
\usepackage{url}

\usepackage{titlesec}
\titleformat{\section}{\normalfont\Large\bfseries}{\thesection}{1em}{}
\titleformat{\subsection}{\normalfont\large\bfseries}{\thesubsection}{1em}{}
\usepackage{environ}
\NewEnviron{abstractblock}{%
  \begin{center}\textbf{Abstract}\end{center}%
  \begin{quote}\BODY\end{quote}%
}

\title{\textbf{Epi-Logic: A Conceptual Framework for Epistemic Runtime Control,\\
Schema Validity Checking, and Controlled Accommodation in Autonomous AI Agents}}
\author{Boris Wetzk\\
\small Independent Researcher, Hamburg, Germany\\
\small \texttt{Boris.Wetzk.Research@web.de}}
\date{Preprint}

\begin{document}
\maketitle

\begin{abstractblock}

Autonomous AI agents are increasingly deployed in areas where wrong
decisions are hard to reverse. This paper examines schema mismatch: the
condition in which an agent operates within an interpretive frame that
no longer applies to the current context. Outputs produced under such a
mismatch can appear internally consistent, linguistically plausible, and
largely factually correct; output-quality metrics alone therefore
capture the underlying loss of validity only partially.

The paper introduces Epi-Logic, a conceptual framework for epistemic
runtime control. It couples the detection of schema dissonance, a
graduated reduction of autonomy, and the auditable switch to a validated
schema. A schema is formalised as a tuple of variable space, expectation
model, validity conditions, axioms, and metadata. The Epi-Score
aggregates seven graded dimensions of epistemic dissonance; the temporal
validity dimension D8, violations of the validity conditions G, and
axiom violations are carried as separate categorical paths that are not
offset against the aggregate.

The architecture rests on a checking asymmetry: formalised validity
conditions can be checked at runtime, whereas the correctness of many
actions is established only ex post. Whether those conditions fully
capture the schema\textquotesingle s real applicability remains an
empirical question. The paper separates two architectural properties, a
conditional result from sequential changepoint detection, and this
empirical remainder. Eight falsifiable propositions with named baselines
and a proof-of-concept design describe the transition to empirical
validation; governance implications are discussed with reference to the
EU AI Act.

\emph{Status: conceptual paper. All propositions are empirically
testable hypotheses, not established results.}


\end{abstractblock}

\section{Introduction}\label{introduction}

\subsection{Background and
Motivation}\label{background-and-motivation}

Autonomous AI agents are increasingly deployed in domains where
decisions are hard to reverse: financial analysis, clinical decision
support, legal research, industrial control. These systems combine large
language models (LLMs) with tools, memory, and decision logic into
architectures that plan, act, and justify their actions across multiple
steps.

Hallucination and out-of-distribution behaviour are two established
reference frames in reliability research; alongside them, agent-specific
failure forms such as goal, constraint, or context drift are being
studied. Schema mismatch cuts across these categories, because the fault
concerns the interpretive frame under which information is weighted and
actions are derived.

Empirical work shows related patterns. Cemri et al. (2025) attribute
32.3 per cent of failures across 1,642 annotated multi-agent execution
traces to inter-agent misalignment; Xie et al. (2026) model error
cascades; Wang, Jin, Cao and Wang (2026) examine epistemic
miscalibration under correct plan execution; and Rath (2026) studies
simulation-based agent drift in financial-analysis settings. These
findings do not establish schema mismatch as a unified class, but they
show the practical relevance of internally consistent misfits.

For the temporal sub-question, Elbadry et al. (2026) show in controlled
experiments that temporal knowledge drift can be distinguished in
representation space from correctness and uncertainty. This motivates
monitoring temporal validity separately.

\subsection{Problem Statement}\label{problem-statement}

The central problem of this paper can be sharpened into two questions.

The first is diagnostic: how can it be recognised that an autonomous AI
agent is operating under a schema that no longer applies to the current
context, and how can this be recognised before the state becomes visible
as measurable output degradation?

The second is architectural: how can a system be designed to translate
such a detection into a proportionate response? Proportionate here means
graduated reduction of autonomy, escalation, or a controlled switch to a
better-suited schema, in place of a blanket halt.

Existing approaches, in particular OOD detection, uncertainty
quantification, and runtime assurance, address parts of the problem. The
combination of explicit schema representation, continuous checking of
its validity conditions, externally anchored temporal control, graduated
autonomy regulation, and documented switching between pre-validated
schemas is provided by none of these approaches.

\subsection{Research Gap}\label{research-gap}

A synthesis of existing approaches points to three gaps that Epi-Logic
sets out to close. They concern the applicability of an explicit
interpretive frame as an object of runtime control, and the response
logic tied to it.

First, a unified response logic is missing. OOD detection and
uncertainty quantification yield signals without a defined course of
action. Runtime assurance systems define intervention points for
constraint violations, and some of them provide graded responses; the
gradual decline in the applicability of an explicit interpretive frame
is not among the conditions they check. The link between signal,
interpretation, and graduated response is absent from existing
architectures.

Second, the temporal validity layer has so far been confined to
individual knowledge elements. Existing work checks whether a fact, a
retrieved version, or a memory state still holds. The follow-on
question, whether the interpretive frame in which these elements are
used still applies to the current action, is not addressed in that work
and is not connected to any response logic.

Third, structured auditability of accommodation is missing, that is, of
the controlled switch of the active schema (formally defined in Section
3.4). Continual learning and schema-switching mechanisms exist; in
regulated domains, traceability of which schema was activated or
deactivated, and on what grounds, is a plausible governance expectation.
Such journaling structures are not consistently included in the
frameworks considered.

\subsection{Aim of the Paper}\label{aim-of-the-paper}

The aim is to propose a formally bounded and empirically testable
framework that treats schema validity as a runtime object,
operationalises dissonance through an Epi-Score, and derives from it
graduated intervention and auditable accommodation. The names Epi-Logic
and Epi-Score refer to episteme, that is, to knowledge and the question
of its standing; the framework poses that question at runtime. The
empirical robustness of the framework remains open and is translated
into testable propositions.

\subsection{Contributions}\label{contributions}

B1, conceptual distinction: systematic consolidation and delimitation of
schema-related misfits as a distinct runtime perspective relative to
hallucination, OOD behaviour, and general forecast deviation, including
a conceptual separation of noise and epistemic dissonance.

B2, schema formalism: definition of the schema as a five-tuple of
variable space, expectation model, validity conditions, axioms, and
metadata, with explicitness as an applicability condition of the
framework.

B3, architectural specification: formalisation of Epi-Logic as a
two-layer control architecture with an operational agent (Layer 1), an
epistemic monitor (Layer 2), schema library, calibration layer, and
intervention logic, grounded in cybernetic control-loop theory and in a
checking asymmetry decomposed into two architectural properties, a
conditional link to changepoint theory, and an empirical remainder.

B4, measurement instrument: specification of the Epi-Score as a
multi-dimensional runtime metric with seven aggregated dimensions, a
veto dimension for temporal validity operating outside the aggregation,
a dynamic threshold, and a weighting architecture that permits
domain-specific calibration.

B5, research agenda: derivation of eight propositions with named
baselines and a structured proof-of-concept design enabling the
transition from conceptual to empirical research.

\subsection{Methodological Approach}\label{methodological-approach}

This is a conceptual paper; it combines synthesis of existing theory
with the construction of a model. The research streams in Section 2 were
selected according to whether they define an object of checking at
runtime: distributional membership, uncertainty, goal conformity, rule
conformity, or the temporal validity of knowledge. The schema formalism
arose from asking which components an interpretive frame needs for the
satisfaction of its validity conditions to be decidable at runtime; the
five components of the tuple constitute the decomposition adopted here,
and the explicitness condition follows from the same requirement. Other
decompositions serving the same purpose are not excluded. Dimensions D1
to D8 were derived from the three causes of mismatch distinguished in
Section 3.1: structural, epistemic, and temporal. The propositions arose
by asking, for each load-bearing architectural assumption, which
observation would refute it.

The statements in this paper have differing status, named at each point
in the text. Definitions fix terms and are not truth-apt. Design
decisions justify the construction and could have been made otherwise.
The paper claims no formal result of its own. The architectural
properties in Section 4.4 follow from the construction and from the
scope of application. The conditional link to changepoint theory imports
external results under named assumptions. The propositions are empirical
hypotheses without evidence.

\subsection{Structure of the Paper}\label{structure-of-the-paper}

Section 2 situates the state of research, Section 3 defines the problem
and scope, Section 4 presents the framework, and Section 5 addresses
operationalisation and validation. Section 6 discusses implications and
limitations, and Section 7 presents the research agenda.

\section{Related Work and Theoretical
Foundations}\label{related-work-and-theoretical-foundations}

\subsection{Research Streams}\label{research-streams}

Research on the reliability and safety of autonomous AI agents can be
grouped into six streams that touch on Epi-Logic without covering it
fully.

Out-of-distribution detection tests whether an input lies outside a
reference distribution. Entropy-based, confidence-based, and more recent
weight-based methods provide established signals for this (Malinin and
Gales 2021; de Mathelin et al. 2025). In Epi-Logic, OOD signals are
possible inputs, but they do not capture schema validity on their own.

Uncertainty quantification estimates predictive uncertainty through
ensembles, Bayesian neural networks, and multi-step agent procedures,
among others (Oh et al. 2026; Han et al. 2024). An internally consistent
but ill-fitting schema can nonetheless coincide with low measured
uncertainty.

Runtime assurance monitors autonomous systems during operation and
intervenes when verifiable safety conditions are violated. MI9 (C. L.
Wang et al. 2025) offers six coordinated mechanisms including drift
detection and graded containment logic; Kaptein et al. (2026) formalise
runtime governance over agent paths; an earlier precursor learns safety
interventions through reinforcement learning (Lazarus et al. 2020). The
older stream of self-adaptive systems had already formulated the
separation of operational logic and adaptation logic as an architectural
principle: MAPE-K control loops of monitor, analyse, plan, and execute
over a shared knowledge base (Kephart and Chess 2003), and explicit
runtime models in the models@run.time approach (Blair et al. 2009).
Epi-Logic adopts this two-part division and gives the analysis stage an
object of checking of its own, the validity of the interpretive frame;
what is added here is a specific object of checking: the applicability
of an explicit interpretive frame to the current situation. Runtime
assurance focuses on safety boundaries and rule-based compliance; the
epistemic question of validity lies outside its scope.

Continual learning and open-world recognition connect novelty detection
with incremental adaptation after deployment (Bendale and Boult 2015;
Kim et al. 2024). The question of when a system may continue to act
autonomously during such adaptation is not treated there as a unified
object of control.

LLM self-correction and evaluator models improve consistency and factual
fidelity through chain-of-thought variants, retrieval-augmented
generation, and external evaluator models. Romanchuk and Bondar (2026)
formalise, under their architectural assumptions, limits of circular
epistemic justification; Quattrociocchi, Capraro and Perc (2025) discuss
the absence of internal metacognition. This work motivates the
structural separation of Layer 1 and Layer 2; why external checking can
offer an advantage is developed in Section 4.4 as the checking
asymmetry.

Context drift and agent memory address related manifestations: goal
drift relative to original instructions (Arike et al. 2025),
memory-induced tool drift (Dabas et al. 2026), constraint drift in
multi-agent systems (Li et al. 2026), and diverging knowledge states
across agents (Rodrigues 2026). This work supplies failure patterns and
measures; Epi-Logic consolidates them under the question of whether an
explicit interpretive frame still applies, and what runtime response
follows from that.

\subsection{Theoretical Foundations}\label{theoretical-foundations}

Epi-Logic draws on four theoretical traditions that together form the
conceptual foundation of the framework.

Model validity and distribution shift. The statistical concept of
distribution shift (Quiñonero-Candela et al. 2009; Liu et al. 2021)
supplies the technical core. A model can minimise its error on known
data and operate in an environment for which it was not calibrated.
Epi-Logic asks less about the magnitude of deviation from the training
distribution than about whether the active interpretive frame still
applies to the current context. This is a question of model validity
rather than of prediction error.

Classical epistemology and action-guiding knowledge. The classical
analysis treats knowledge as justified true belief; it has long been
regarded as necessary but not sufficient. For the question pursued here,
the justification condition is the relevant one: an agent can hold an
output with high confidence while its basis is no longer justified at
the moment of action. Whether beliefs in the epistemological sense are
to be ascribed to a model remains open for this argument. Time-indexed
truth is an established concept in epistemology; a statement is true at
a time t. If a proposition no longer holds at the time of action, it
already fails this time-indexed truth condition; Epi-Logic does not
thereby claim an additional condition of knowledge. For the static
ascription of knowledge, this indexing suffices. An acting agent faces
an additional operational problem: its knowledge base is calibrated to a
time t\_k, while it acts at a time t\_a, and the interval between the
two is not accessible to it. Validity denotes the operational condition
that knowledge calibrated or validated at t\_k still applies at t\_a,
not a separate condition of knowledge alongside truth, belief, and
justification. Elapsed time alone does not invalidate it; what matters
is whether a relevant supersession, change of context, or change of
jurisdiction has occurred in between. A statement can be true and
justified within a model\textquotesingle s knowledge base and still fail
to hold for the current moment of action for one of those reasons.

Cybernetics and control-loop theory. The separation of acting and
monitoring subsystems is an established architectural principle: the
adaptation logic is implemented as a separate control loop above the
operational system (Kephart and Chess 2003; Blair et al. 2009). Eslami
and Yu (2026) formalise, for agentic systems, that agency can be
modelled as hierarchical runtime decision authority within a closed
loop, and that stability depends on the coupling between adaptation,
switching, and delay. Epi-Logic applies this separation to a distinct
object of checking: the monitor assesses the validity of the active
schema without solving the task of Layer 1. The epistemic independence
of the monitor is a condition for its ability to detect classes of
failure that remain structurally invisible to the agent.

Reference-based calibration. Epi-Logic interprets dissonance relative to
calibrated reference distributions. Statistical process control provides
an established underlying principle: deviation is assessed against a
defined state of normal operation (Shewhart 1931). Depending on the
dimension, this benchmark may derive from an external population, a
historical reference state, an experimental baseline, or a normative
bound.

\subsection{Limits of Existing
Approaches}\label{limits-of-existing-approaches}

The six research streams address substantial parts of the problem, but
not the same combination of schema validity checking, autonomy
regulation, and auditable schema switching. The closest overlap concerns
temporal knowledge validity.

Yadav (2026), Chao et al. (2026), and Elbadry et al. (2026) test whether
individual facts, versions, or memory states still hold. Epi-Logic
treats such findings as possible D8 signals but distinguishes them from
the broader question of whether the interpretive frame as a whole still
applies to the current situation. Current facts do not guarantee a valid
framing; conversely, a single outdated element does not automatically
invalidate the entire schema.

\section{Problem Definition and Conceptual
Scope}\label{problem-definition-and-conceptual-scope}

\subsection{Definition of the Core
Problem}\label{definition-of-the-core-problem}

An autonomous AI agent operates within an active interpretive frame that
determines how incoming information is processed, weighted, and turned
into output. This frame is termed a schema and is formally defined in
Section 3.4.

Schema mismatch denotes the condition in which the
agent\textquotesingle s active schema no longer fits the current
situation. The resulting outputs are typically internally consistent and
linguistically sound; the problem lies in the incongruence between
schema and context. This unremarkable surface form is characteristic of
the failure class rather than a defining condition of it. The term
schema is not to be understood here in the sense of data-structure
validation, such as a JSON schema checking the shape of a record. What
is meant is the interpretive frame under which an agent construes and
weights information.

This failure form is to be distinguished from hallucination: with
hallucination, the content is typically false or unsupported; with
schema mismatch, the facts used may be correct while they are weighted
or interpreted within an ill-fitting frame.

Three subtypes can be distinguished. Structural mismatch is present when
the schema does not fit the task context semantically. Epistemic
mismatch is present when the schema contains insufficiently grounded or
miscalibrated assumptions. Temporal mismatch is present when the schema
was correct but has lost its applicability through changed
circumstances.

These three subtypes describe what is wrong with the schema. An
independent distinction concerns the origin of the deviation.
Structural, epistemic, and temporal mismatches arise when the world or
the task shifts relative to a schema that initially fitted. Technical
mismatch arises when the schema is altered by the
agent\textquotesingle s own processing infrastructure, irrespective of
whether the external world has moved. One documented example is the loss
of accumulated context through compaction, the iterative rewriting and
summarising of long conversations to stay within the context window,
which erodes detail across repeated cycles (Zhang et al. 2025).

\subsection{Applicability
Conditions}\label{applicability-conditions}

Epi-Logic applies to systems that satisfy five conditions.

The first condition is schematicity: the agent operates in a context in
which different interpretive frames with different validity conditions
can be distinguished.

The second condition is explicitness: the interpretive frame must be
specifiable as an artefact in the sense of Section 3.4. For LLM agents
this requires making task frames, tool policies, the scoping of the
knowledge base, and constraints explicit in the orchestration. Epi-Logic
does not claim to reconstruct latent interpretive structures in model
weights; systems without a sufficiently explicit frame lie outside its
scope.

The third condition is consequentiality: the agent\textquotesingle s
outputs have a bearing on action.

The fourth condition is ex-post verifiability: the target domain
contains reference points against which schema mismatch can be
established after the fact.

The fifth condition is independence of the observation stream: at least
one signal source relevant to D6, D8, or the validity check must be
neither produced nor mediated by the monitored agent. Without additional
independent evidence, further internal communication does not create an
informational advantage regarding the state of the world (Ao, Gao and
Simchi-Levi 2026). A shared drift of the system relative to the external
world is therefore not reliably addressable by internal exchange alone.
Section 4.4 treats the resulting informational asymmetry in more detail.

\subsection{Underlying Assumptions}\label{underlying-assumptions}

The framework rests on five assumptions.

Assumption 1: schema mismatch is measurable before it becomes visible as
output degradation.

Assumption 2: an external monitor with structurally different signals
detects schema mismatch more reliably than self-evaluation.

Assumption 3: schema switching with documented validation reduces harm
after regime changes compared with continued operation under the old
schema.

Assumption 4: temporal knowledge validity is a distinct dimension of
failure that is not fully covered by existing methods.

Assumption 5 (checking asymmetry): checking a schema\textquotesingle s
explicit validity conditions can, with equal resources, be performed
more reliably than the task within that schema can be solved. Section
4.4 decomposes this assumption into two architectural properties, a
conditional link to changepoint theory, and an empirically open
remainder; it is that remainder which is testable in the sense of
Proposition P5b.

\subsection{Terminology and Schema
Formalism}\label{terminology-and-schema-formalism}

Schema. A schema is an explicit design artefact of the agent
orchestration, defined as a five-tuple

S = (V, E, G, A, M)

with the following components:

V (variable space): the finite set of quantities the schema observes and
interprets, including their value ranges and measurement sources.

E (expectation model): a function that generates, from the current state
of the variable space, expectations about future or concurrent
observations. E may be formulated parametrically (explicit rules,
assumed relationships) or be learned (a representation space describing
normal behaviour in this context). Both forms may be combined.

G (validity conditions): a finite, machine-checkable set of conditions
over V under which the schema applies, including the
schema\textquotesingle s own tolerance bands for expected deviation. G
is formulated so as to be falsifiable: for each condition, it is
specified which observation would violate it.

A (axioms): hard boundary conditions whose violation justifies
intervention independently of all other signals, such as safety, legal,
or plausibility bounds.

M (metadata): version, provenance, validation status, temporal
calibration of the knowledge state, applicable alternative schemas, a
declared distance measure for parameter adjustments, and the state of
the drift budget (cumulative parameter distance since the last
validation, and the calibrated budget limit).

Schema identity is determined by V, E and G: two schemas count as
different if they differ in the variable space V, in the structure of E,
or in G. If they differ only in parameter values of E within identical V
and G, they are parametrisations of the same schema. A and M do not bear
on identity; axioms constrain the permissibility of actions across
schemas, and metadata record version and provenance. This identity rule
separates schema switching (accommodation) from parameter adjustment
(assimilation with adjustment).

The separation via identical validity conditions has a borderline case:
many small parameter adjustments can cumulatively approach structural
change. M therefore carries a drift budget B that accumulates the
distance between successive parametrisations since the last validation.
The distance measure is to be fixed per schema type. If B exceeds a
calibrated limit, the current parametrisation is revalidated in the
sandbox; on success it becomes the new reference state and B is reset,
otherwise schema selection applies. The budget thereby bounds drift on
the schema side, while D3 captures shifts of the context.

Schemas exist on two levels. Domain schemas describe the shared
interpretive frame of a domain and are maintained centrally. Instance
schemas are customer-specific adaptation layers built on top of domain
schemas.

Schema mismatch denotes the incongruence between the active schema and
the current context, operationally: the violation of conditions in G, or
the persistent divergence between the expectations from E and the
observations over V. Violations of the axioms in A are treated
separately: an agent can select a prohibited or safety-critical action
while operating under a fitting schema. Such violations justify
intervention on safety and compliance grounds and are carried as a
categorical path of their own (Section 5.2).

Epistemic dissonance denotes the measurable state in which observation,
expectation, and validity conditions of a schema no longer cohere.
Section 3.5 delimits dissonance from noise.

Epi-Score denotes the composite runtime metric that aggregates epistemic
dissonance across several dimensions (Section 5.2).

Accommodation denotes the process by which an agent switches its active
schema. The term follows Piaget\textquotesingle s distinction between
assimilation and accommodation (Piaget, 1952). Triggered by the
intervention logic, the switch is tested against the acceptance criteria
named in Section 4.3 and accompanied by sandbox validation and audit
documentation.

Layer 1 denotes the operational agent; Layer 2 denotes the epistemic
monitor.

\subsection{Noise and Dissonance}\label{noise-and-dissonance}

Every schema produces expected deviation. An expectation model admitting
no dispersion at all would be unusable in real environments. Expected
deviation (noise) and evidence against the continued applicability of
the schema (dissonance) can be separated conceptually.

Noise is a deviation that is compatible with the variability expected
under the active schema and does not by itself provide sufficient
evidence against the applicability of the schema.

Dissonance is measurable evidence that observation, expectation, and
validity conditions are no longer compatible with the continued
applicability of the active schema. Violations of the axioms in A are
not part of this notion: they concern the permissibility of an action
rather than the applicability of the frame, and they are carried as a
categorical path of their own (Section 5.2).

Both terms describe states, not a detection procedure. Which signal
patterns indicate dissonance in practice is an empirical question.
Candidates include exceedance of the tolerance bands in G, correlated
occurrence across structurally independent dimensions, persistence
across the schema-defined observation window, and abrupt single findings
of large amplitude; their combination and thresholds are to be
calibrated per domain, since base rate, measurement frequency, and cost
of harm affect them. Section 5.2 specifies the decision rule used,
Section 5.3 its validation.

\section{The Conceptual Framework}\label{the-conceptual-framework}

\subsection{Overview}\label{overview}

Epi-Logic is a two-layer control architecture for epistemic runtime
control in autonomous AI agents. The architecture resolves a basic
tension in autonomous systems: they must act quickly and efficiently,
yet should also recognise when action rests on an invalid basis.
Epi-Logic resolves this tension through structural separation: an
operational agent (Layer 1) acts efficiently within an active schema; an
epistemic monitor (Layer 2) assesses the validity of that schema in
parallel, without intervening in task execution.

The calibration layer normalises the heterogeneous signals against a
domain-specific reference group. The Epi-Score aggregates D1 to D7 into
a composite runtime metric for epistemic dissonance, and the
intervention logic translates the monitored evidence into graduated
responses. The cognitive journal records every intervention step in an
auditable manner.

\begin{figure}[htbp]\centering\includegraphics[width=\textwidth]{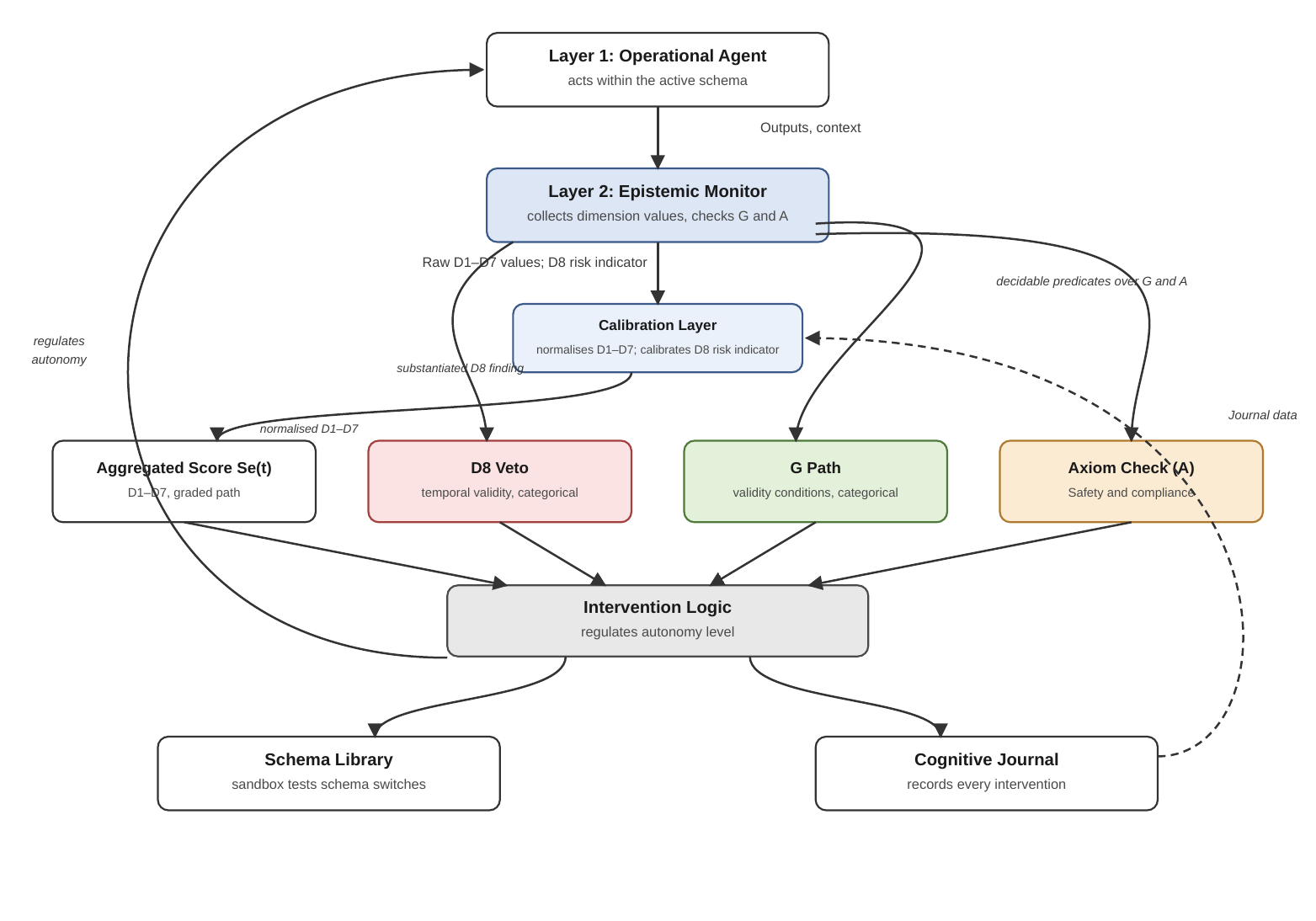}
\caption{The Epi-Logic control loop. Layer 1 acts within the
active schema; Layer 2 collects the dimension values and checks validity
conditions G and axioms A as decidable predicates. The calibration layer
normalises D1 to D7 for the aggregated score and calibrates only the
graded D8 risk indicator; a substantiated D8 finding is a categorical
veto and is routed directly from Layer 2. Four paths reach the
intervention logic separately: the aggregated score Se(t) over the
graded dimensions D1 to D7, the categorical D8 veto for temporal
validity, the G path for violations of the validity conditions, and the
axiom check over A for safety and compliance. The intervention logic
regulates the autonomy of Layer 1, tests schema switches through the
sandbox of the schema library, and documents every intervention in the
cognitive journal. Journal data feed back into the calibration layer.}
\end{figure}

\subsection{Core Components}\label{core-components}

The architecture comprises five core components.

Layer 1: the operational agent. Layer 1 is the performance-oriented
system. It may be an LLM agent, a trading model, a clinical classifier,
or a hybrid system. Layer 1 acts within an active schema using heuristic
and probabilistic methods and is optimised for task performance. It may
carry internal checks of its own, such as self-critique or
self-consistency procedures. Layer 2 is added to these as a structurally
separate instance with a different object of checking and access to
signals that are unavailable to the agent.

Layer 2: the epistemic monitor. Layer 2 is a structurally separate
control instance. Alongside the outputs, it observes the relation
between inputs, internal expectations, observed consequences,
uncertainty values, and the validity conditions of the active schema.
The monitor assesses whether the frame in which Layer 1 solves the task
still applies to the current context. Its structural separation and its
access to additional, partly agent-independent signals are conditions of
the architecture; Section 4.4 justifies the checking advantage that may
follow from them.

Schema library. The schema library manages validated schemas in the
format of Section 3.4, their versions, and their applicability
relations, in particular the alternatives to be tested when an active
schema becomes invalid.

Calibration layer. It normalises dimension values against calibrated
reference distributions, supplies stability information for dynamic
thresholds, and is updated by confirmed or refuted alarms. An external
reference population is a preferred but not the only instantiation. So
as not to absorb creeping or collective drift into the baseline,
reference states are versioned and periodically checked against
historical or otherwise validated contrast cases. The remaining
dependence on reference quality is a limitation (Section 6.5).

Intervention logic and cognitive journal. The intervention logic
translates the result of the score evaluation into a graduated response.
It distinguishes four states: normal operation with full autonomy,
warning mode with reduced confidence, restricted mode requiring human
approval for consequential actions, and safe state with autonomous
action suspended. Every step is recorded in the cognitive journal:
triggering signal, active schema, alternatives tested, affected axioms,
timestamp.

\subsection{Process Logic}\label{process-logic}

The operation of Epi-Logic can be described in six states that form a
control loop with possible returns to earlier states.

Stable assimilation. Layer 1 processes data within the active schema;
the Epi-Score stays within the expected corridor. Layer 2 runs at
reduced checking intensity.

Rising dissonance. Several signals indicate declining explanatory power;
the criteria from Section 3.5 are partly met. The Epi-Score rises but
stays below the intervention threshold $\tau$ (defined in Section 5.2). The
system reduces confidence values and slows consequential actions.

Trigger and branching. Four events reach the intervention logic on
separate paths, and they lead to different responses. A violation of an
axiom from A triggers a safety and compliance intervention: the action
is blocked or referred for approval, without any implication that the
active schema has become invalid. A violation of a condition from G
establishes, under the definition in Section 3.4, that the active schema
is invalid, and the process continues directly with schema selection. A
substantiated D8 finding triggers at least warning mode and a
revalidation of the affected knowledge base; a single superseded element
does not by itself invalidate the schema, unless it meets the escalation
criteria from Section 5.1, in which case the same G path applies. An
Epi-Score above $\tau$, or a persistent rise on the trend path, indicates
dissonance across several dimensions and likewise raises the question of
schema validity; this case also continues with schema selection.

Schema selection. Layer 2 tests alternative schemas in a sandbox against
five acceptance criteria: the candidate satisfies G and all relevant
axioms from A; dissonance decreases over a pre-defined observation
window; task performance does not fall below a specified minimum; the
new state remains within calibration and drift bounds; rollback option,
approval status, and reason for the switch are documented in the
cognitive journal. If the library yields no candidate that meets all
five criteria, the switch does not take place; Epi-Logic then reduces
autonomy up to the safe state and escalates to human review.

Adjustment of the frame. Parameter adjustments remain within the same
schema and draw on the drift budget; selecting a different validated
schema, or synthesising a new one, constitutes accommodation. A first
implementation is confined to parameter adjustment and selection;
synthesis remains a later research goal.

Re-equilibration and audit. The cognitive journal records the complete
transition. Usage behaviour and corrections feed back into schema
improvement and into the calibration layer.

\subsection{The Checking Asymmetry: Architectural Properties,
Conditional Link to Changepoint Theory, and Empirical
Remainder}\label{the-checking-asymmetry-architectural-properties-conditional-link-to-changepoint-theory-and-empirical-remainder}

Romanchuk and Bondar (2026) motivate, under their architectural
assumptions, the separation of self-evaluation and external checking.
What remains open is why an external monitor should be able to perform
checks more reliably, given that it too requires an interpretive frame
of its own. Epi-Logic answers this question on three levels: through two
architectural properties, a conditional result from sequential
changepoint detection, and an empirical remainder.

First architectural property: asymmetry of runtime checkability. Let S =
(V, E, G, A, M) be a schema whose validity conditions G are given as a
finite set of n decidable predicates over V with bounded evaluation
cost. Where axioms in A are specified in the same machine-checkable
form, they are evaluated on the same basis, though they serve the
separate safety and compliance path. Their complete checking then
terminates per observation cycle with cost O(n). The correctness of
Layer 1\textquotesingle s output cannot necessarily be established at
the same moment and under the fourth applicability condition, because
the governing reference point becomes available only ex post. What is
decidable here is solely the satisfaction of the formalised predicates,
not the validity of the schema against the world; whether G covers the
relevant failure modes remains empirically open. The asymmetry is thus a
property of the chosen scope. For latent semantic properties that are
not made explicit, no guarantee of decidability is claimed (Rice 1953).

Second architectural property: informational asymmetry. By construction,
the monitor has a signal set that extends the agent\textquotesingle s,
for instance by sampling statistics, external freshness and validity
data, metadata of the knowledge base, and reference distributions. An
optimal detector with a proper superset of information can at least
reproduce the rules available on the smaller information set; the
attainable upper bound of its detection performance is therefore not
lowered by additional information. Whether a concrete implementation
exploits this potential advantage is a matter for empirical testing. Ao,
Gao and Simchi-Levi (2026) support the underlying separation: purely
internal communication generates no additional informational value about
the state of the world at a fixed exogenous information level.

This difference in information sets also determines the
monitor\textquotesingle s status with respect to the decision-theoretic
result of Ao, Gao and Simchi-Levi (2026). Their theorem separates
checking stages that reprocess the same information level and do not
raise the decision ceiling from checking stages that add a non-redundant
exogenous signal and can raise it. The Layer 2 monitor is constructed so
as to draw on quantities outside the information level of the monitored
agent. Whether these signals are non-redundant and informative for
schema mismatch in a given domain does not follow from the construction
and is an empirical question.

Conditional result. The reference-normalised values D1 to D7 form an
observation stream. T1: under a valid schema, the observation stream
follows a pre-change distribution that is treated as known for the
purposes of the imported result and is estimated from the calibration
layer in practice. T2: schema invalidity produces a post-change
distribution, likewise treated as specified, with positive divergence
from the pre-change distribution. T3: the change carries through to the
monitored detection statistic; for the aggregate path, a shift of the
expected value of Se by at least $\delta > 0$ is assumed. T4:
within a regime, the observations are conditionally independent. T3 does
not follow from T2, because changes in distributional shape or
countervailing dimensions can remain invisible in the aggregate. Under
T1, T2 and T4, classical results on detection delay and mean time to
false alarm hold for likelihood-ratio-based CUSUM (Page 1954; Lorden
1971; Moustakides 1986). For estimated, unknown, or dependent cases,
extensions exist that require their own assumptions and are not covered
by the guarantee cited here. The score recursion used in Section 5.2 is
a mean-shift detector and inherits these optimality statements only
insofar as it corresponds to the likelihood-ratio statistic of the
chosen distributional model; with an unknown post-change distribution,
adaptive or GLR variants are required. Evaluation proceeds in sample
time, which matters under variable agent cadence compared with pure
wall-clock calibration (Modgil 2026). Romanchuk and Bondar motivate,
under their architectural assumptions, limits of circular epistemic
justification. Whether self-evaluation is worse at accessing the signal
required for detection does not follow from this and remains the subject
of P5b.

The empirical remainder consists of three correspondence questions
between specification and world: does G cover the relevant failure
modes? Does schema invalidity produce a sufficiently strong shift in the
monitored statistic? Does the observation stream satisfy T4
sufficiently? None of these questions can be answered from the
architecture alone. They are to be tested empirically per domain. P5a
addresses the second of them, the observable shift; P5b additionally
tests the comparative monitoring advantage; the coverage of G and the
dependence structure assumed in T4 remain separate domain-specific
validation conditions. A negative result also localises where adjustment
is needed: in the schema specification, the set of dimensions, the
detection statistic, or the monitor implementation.

The decomposition thereby makes falsification localisable: an
unobservable shift leaves the imported changepoint result untouched and
bears on the assumption that the chosen dimension and detection space
capture the relevant mismatch.

The regress of checking the monitor is likewise not resolved logically.
Epi-Logic sets an operational governance boundary: alarms and monitor
changes are validated ex post, undergo independent quality assurance,
and, for critical decisions, are subject to human accountability. The
mean time to false alarm can serve as a calibration quantity to steer
the checking load; missed shifts remain unaffected by this and must be
captured through empirical false-negative analysis.

\subsection{Framework Propositions}\label{framework-propositions}

The following eight propositions derive from the framework. They are
formulated as empirically testable hypotheses; for each, the baseline
procedure against which it is tested is named.

P1, temporal lead of non-temporal schema mismatch over output
degradation. Agents operating under a structurally or epistemically
miscalibrated schema show a qualified alarm from dimensions D1 to D7
ahead of the first demonstrable degradation of established
output-quality metrics, with subtype-specific dimension profiles,
compared with agents under a contextually valid schema. The temporal
case is excluded and is the subject of P3a, in which D1 to D7 may show
no obvious anomaly. Testing uses cases with a documented point of
change: the time of the first qualified alarm and the time of the first
demonstrable output degradation are recorded separately; the target
quantity is the lead time between the two events. Nested comparison, to
control for the measurement overlap of D5 and D6 with the output
baselines: first established output-quality metrics alone, then D1 to D4
together with D7, then the full profile D1 to D7. Baseline: established
output-quality metrics (faithfulness, groundedness, answer relevance) as
implemented in widely used RAG evaluation suites.

P2, confidence asymmetry as a leading signal. In multi-step agent
workflows, D7 (confidence asymmetry) rises before measurable output
degradation. Operationally, a lead is defined as at least one completed
tool or action step between the rise in D7 and demonstrable performance
decline, measured in the recorded action sequence. Choosing an external
sequence unit rather than a model-internal reasoning step keeps the
measure comparable across models. Baseline: simple confidence and
entropy measures of the token distribution without repeated sampling.

P3a, temporal validity as a distinct dimension of failure. LLM agents
accessing knowledge bases that contain superseded information produce
substantiated D8 findings with predictive value beyond D1 to D7. Testing
proceeds as an incremental contribution: a binary indicator for a
substantiated D8 finding is added to a model built from D1 to D7, and
the analysis tests whether the hit rate on documented supersession cases
increases. The ground-truth annotation of supersession is kept separate
from the procedure that produces D8: D8 must determine current authority
itself, from realistically available sources and metadata, without
access to the annotation. Baseline: standard hallucination detection
through self-consistency checking (repeated sampling with consistency
comparison, for instance following the SelfCheckGPT principle, Manakul,
Liusie and Gales 2023), and ensemble-based uncertainty quantification.
The proposition is confirmed if D8 detects cases of superseded
information for which neither baseline shows an obvious anomaly.

P3b, assignment of D8 findings to schema-wide validity. The escalation
rule from Section 5.1 reliably assigns substantiated D8 findings to a
violation, or non-violation, of the validity conditions G. Testing uses
an annotated case corpus in which, for each substantiated D8 finding, it
is recorded independently of the rule whether the finding calls the
applicability of the schema as a whole into question or remains confined
to the individual knowledge element. The target quantity is the
agreement of the rule-based assignment with this annotation, measured as
F1. Baseline: a fixed threshold based only on the age or number of
superseded elements, without the remaining escalation criteria from
Section 5.1.

P4, schema accommodation improves recovery. Agents with schema
accommodation logic show faster recovery and lower cumulative error
after regime changes. Baselines: first, continued operation under the
original schema; second, an unstructured restart of the agent without
schema-switching logic; third, an established adaptive procedure without
explicit schema representation, such as rolling refitting of the model
or regime-based routing between pre-trained models.

P5a, detectability. In documented schema-mismatch episodes that satisfy
T1, T2, and T4 of the conditional link to changepoint theory (Section
4.4), the statistic monitored by Layer 2 exhibits the shift specified in
T3 strongly enough for sequential detection. Testing determines whether
T3 holds in the target domain. If the required shift does not occur, T3
is not satisfied and the set of dimensions or the monitored statistic
must be revised.

P5b, incremental external information value. An architecturally
independent epistemic monitor with agent-external signals achieves lower
detection delay at a fixed average run length to false alarm and higher
recall at a fixed false-positive rate than both the same detector
without those signals and the self-evaluation baselines, particularly in
internally consistent mismatch cases. Endpoints: detection delay at a
fixed average run length to false alarm, recall at a fixed
false-positive rate, ablations of individual exogenous signals.
Baselines: self-consistency checking, self-critique prompting of the
agent on its own output, and LLM-as-judge with the same base model as
Layer 1, each with the same compute budget as the monitor. The test
separates the sources of effect through ablation: the monitor is
additionally tested without the agent-external signals and with a
simpler detector over the same signals. If P5a\textquotesingle s
required shift holds but the detection gain fails to appear, the cause
may lie in threshold calibration, the dependency structure, a statistic
unsuited to the domain, or the monitor implementation.

P6, dimension profile enables differential intervention. The dimension
profile of the Epi-Score classifies the likely cause of schema mismatch
with sufficient discriminative power to justify different intervention
strategies. Two target quantities are tested separately: first, cause
classification, measured as macro-averaged F1 classification quality
across the documented mismatch types against the annotated cause;
second, intervention utility, measured as the reduction of unnecessary
autonomy restrictions relative to the baseline at an equal detection
rate. Baseline: an aggregated overall score without a dimension profile
on an identical signal basis.

\section{Operationalisation and Measurement
Implications}\label{operationalisation-and-measurement-implications}

\subsection{Dimensions as a Proposed
Taxonomy}\label{dimensions-as-a-proposed-taxonomy}

Translating schema mismatch into measurable quantities follows a
generative principle: each cause category of schema mismatch gives rise
to its own measurement dimensions. The current eight dimensions are the
provisional result of a systematic analysis of three cause categories:
structural misfit (D1 to D4), epistemic weakness (D5 to D7), and
temporal invalidity (D8).

The dimensions are proposed as an agent-type-independent taxonomy; their
object of measurement is conceptually the same for every agent, whether
this assumption holds empirically is open. What depends on agent type is
the instantiation: the concrete quantity by which the dimension is
measured in a given system. For LLM agents the instantiations are
linguistic and statistical in nature; for predictive agents (trading
systems, robotic controllers) they are numerical and physical. Table 1
places both instantiations side by side. The set of dimensions is a
construction, not an empirically derived structure: whether D1 and D2
are sufficiently discriminant, whether the two instantiations of D6
capture the same latent construct, and whether the sampling stability in
D7 may be interpreted as meta-uncertainty are open questions of
construct validation. P6 tests only the classification quality and
intervention utility of the cause assignment; convergent and
discriminant validity, robustness to alternative operationalisations,
functional comparability across agent types, and profile stability
across domains remain separate checks not covered in this round. Raw
signals such as forecast deviation, semantic inconsistency,
meta-uncertainty, and context indication are subsumed by this scheme:
they are instantiations of the dimensions for predictive agents and do
not form a separate level of aggregation.

\emph{Table 1: The eight channels of the monitoring profile, seven of
them in the aggregated Epi-Score and one separate D8 veto channel. For
each dimension, the object of measurement proposed as
agent-type-independent is given, together with the direction and sign of
the transformation prior to aggregation, one instantiation for LLM
agents and one for predictive agents, and the availability of the
corresponding measurement technique. D1 to D4 capture structural misfit,
D5 to D7 epistemic weakness, D8 temporal invalidity. The final column
indicates whether established measurement procedures exist for the
instantiation; it assesses neither the quality of the measurement nor
the validity of the dimension.}

\begin{longtable}[]{@{}
  >{\raggedright\arraybackslash}p{(\columnwidth - 10\tabcolsep) * \real{0.1701}}
  >{\raggedright\arraybackslash}p{(\columnwidth - 10\tabcolsep) * \real{0.1094}}
  >{\raggedright\arraybackslash}p{(\columnwidth - 10\tabcolsep) * \real{0.1823}}
  >{\raggedright\arraybackslash}p{(\columnwidth - 10\tabcolsep) * \real{0.1762}}
  >{\raggedright\arraybackslash}p{(\columnwidth - 10\tabcolsep) * \real{0.1944}}
  >{\raggedright\arraybackslash}p{(\columnwidth - 10\tabcolsep) * \real{0.1677}}@{}}
\toprule\noalign{}
\begin{minipage}[b]{\linewidth}\raggedright
\textbf{Dimension}
\end{minipage} & \begin{minipage}[b]{\linewidth}\raggedright
\textbf{Direction (s)}
\end{minipage} & \begin{minipage}[b]{\linewidth}\raggedright
\textbf{Object of measurement (agnostic)}
\end{minipage} & \begin{minipage}[b]{\linewidth}\raggedright
\textbf{Instantiation, LLM agent}
\end{minipage} & \begin{minipage}[b]{\linewidth}\raggedright
\textbf{Instantiation, predictive agent}
\end{minipage} & \begin{minipage}[b]{\linewidth}\raggedright
\textbf{Availability of measurement technique}
\end{minipage} \\
\midrule\noalign{}
\endhead
\bottomrule\noalign{}
\endlastfoot
D1 Semantic distance & high = mismatch, s = +1 & Distance of the current
situation from the schema\textquotesingle s core of applicability &
Cosine distance (1 - cosine similarity) of the query to the schema
centroid in embedding space & Distance of the state vector from the
calibrated operating range & established \\
D2 Terminological coherence & high = fit, s = -1 & Agreement of the
situation description with the schema vocabulary & TF-IDF comparison
with core vocabulary & Agreement of active signal sources with the
variable space V & established \\
D3 Contextual drift & high = mismatch, s = +1 & Cumulative shift of the
context over time & Sequential embedding comparison across turns &
Rolling shift of the state distribution across time windows &
established \\
D4 Schema-boundary indicators & high = mismatch, s = +1 & Proximity of
the situation to the edges of G & Trained classifier for boundary
signals (negations, exceptions) & Proximity to, or signed exceedance of,
the tolerance limits in G & partly established \\
D5 Epistemic quality & high = fit, s = -1 & Justification structure of
the agent output; consistency of the reasoning chain & Rule-based
checking, NLI over reasoning chains & Consistency of model assumptions
with the justification & partly established \\
D6 Query-output alignment & high = fit, s = -1 & Congruence between
expectation and delivered performance & Natural language inference
between query and answer & Forecast agreement, operationalised as
inverse divergence between expectation from E and observation &
established \\
D7 Confidence asymmetry & high = mismatch, s = +1 & Instability of agent
behaviour under repetition & Sampling variance under temperature-based
repeated inference & Dispersion of ensemble forecasts; calibration
deviation & to be developed \\
D8 Temporal validity & categorical, outside the sum & Validity of the
knowledge and calibration base at the time of action & External
freshness and validity check, source consistency, metadata & Age of the
calibration data relative to structural-break detection & partly
established \\
\end{longtable}

D8 occupies a special position. Whereas D1 to D7 measure properties of
the output, the process, or the context, D8 measures the temporal
standing of the knowledge and calibration base at the time of action. D8
carries two kinds of finding with different status. A substantiated
finding is present when a piece of information drawn on for the action
has demonstrably been superseded by a newer one; it documents a loss of
validity that obtains even without measurable deviation from
expectation, is categorically different from graded dissonance, and
triggers the veto. A risk indicator is present when the age of the
calibration base alone exceeds a threshold; it justifies increased
checking intensity and does not replace the demonstration. Only the
substantiated finding is carried as a veto signal, and D8 therefore
remains outside the score sum. The applicability of the entire
interpretive frame does not follow from D8 alone; it emerges from the
interplay of the aggregate path, the D8 veto path, the G path, and the
axiom path. Whether a substantiated D8 finding additionally triggers the
G path is governed by the following escalation rule.

Escalation rule. A substantiated D8 finding violates a condition from G
if it meets one or more of the following criteria: the action relevance
of the affected element, its standing within the
schema\textquotesingle s dependencies as a load-bearing rather than a
peripheral assumption, the number of prerequisites affected by the
supersession, and the reach of the supersession event beyond the
individual case. If the finding meets none of these criteria, warning
mode and revalidation of the affected knowledge base apply; the schema
as a whole continues to be considered applicable. Proposition P3b in
Section 4.5 tests whether this rule assigns D8 findings reliably.

\subsection{Score Aggregation, Veto Logic, and
Threshold}\label{score-aggregation-veto-logic-and-threshold}

Before aggregation, each raw dimension is converted into a
dissonance-oriented quantity by a dimension-specific sign
transformation: $d_i(t) = s_i \cdot D_i^{\mathrm{raw}}(t)$, with $s_i \in \{+1, -1\}$. Low values
of di are schema-typical; high values are evidence against the
schema\textquotesingle s applicability. The Epi-Score aggregates the
reference-group-normalised values d1 to d7 as a weighted sum:

\[ Se(t) = \sum_{i=1}^{7} w_i \cdot \mathrm{norm}\big(d_i(t)\big), \quad i = 1, \dots, 7 \]

Here norm($\cdot$) denotes normalisation of the transformed value against the
reference distribution of the applicable domain segment (calibration
layer, Section 4.2). The weights wi are non-negative and normalised to
sum to one. A domain-specific default weight vector forms the starting
point; customer-side adjustments remain within defined corridors while
preserving these constraints. The signs si are fixed per dimension and
listed in Table 1: D1, D3, D4, and D7 enter with si = +1, since their
raw values already point toward mismatch; D2, D5, and D6 enter with si =
-1, since their raw values are inversely oriented and measure fit rather
than dissonance.

D8 and G operate outside this sum. The intervention level results from
the interplay of the aggregate path, the D8 veto path, the G path, and
the axiom path.

The veto rule is as follows: a substantiated D8 finding is itself the
veto event: it activates at least warning mode, irrespective of the
value of Se. The risk indicator of D8 is graded and affects the checking
intensity, without triggering the veto. Violations of axioms in A and
violations of conditions in G do not enter the score. Both are evaluated
by the monitor at runtime as decidable predicates in the sense of
Section 4.4, and they justify intervention directly: an axiom violation
triggers the safety and compliance intervention, a G violation triggers
schema selection under Section 4.3. For G findings, substantiated D8
findings, and axiom violations, compensation by other dimensions cannot
be justified on substantive grounds: a documented loss of validity does
not become less relevant because other dimensions show no obvious
anomaly. Carrying them inside the weighted sum would make their effect
depend on the remaining values. The separation of these paths is
therefore a design decision, taken on these grounds rather than derived
from a formal result.

The threshold $\tau$ is not static. It is formed from a dynamic component
(historical stability of the domain segment from the calibration layer,
current uncertainty, asymmetric costs of false positives and false
negatives) and a structural component (the domain\textquotesingle s risk
of harm).

The decision is based on more than the instantaneous value alone. The
level path compares Se(t) with $\tau(t)$; the trend path carries a sequential
statistic on the CUSUM principle: $C(t) = \max\!\big(0,\, C(t-1) + Se(t) - \mu_0 - k\big)$, where $\mu_0$ denotes the calibrated reference level and k a tolerance
constant. Persistent positive mean shifts can accumulate in this way
even while Se(t) stays below $\tau$. The recursion is the heuristic
mean-shift instantiation of the changepoint approach delimited in
Section 4.4 and claims no general CUSUM optimality outside the
conditions named there.

$\mu_0$ is re-estimated only after validated accommodation, explicit
recalibration, or a drift audit; C(t) is reset at the same time. A
continuously trailing baseline is excluded, because it could normalise
away the creeping shift that the trend path is meant to capture.

\subsection{Validation Strategy}\label{validation-strategy}

The validation of Epi-Logic follows a three-stage approach.

Stage 1, shadow monitoring: the monitor observes passively and records
hypothetical alarms for calibration.

Stage 2, logic guard: the monitor may trigger warnings, confidence
reduction, or human approval for consequential actions.

Stage 3, schema hopping: the system selects among human-approved,
validated schemas; new schemas are not generated autonomously at this
stage.

The primary validation domain for Stage 1 is an LLM agent with a
controlled knowledge base in which outdated and current documents are
deliberately mixed. The ground truth is known and documented. This
constellation tests P3a directly against the named baselines and can be
extended with the escalation annotation needed for P3b. P1 is tested on
a controlled structural or epistemic mismatch with a documented point of
change, in which the task context is shifted relative to a fixed schema;
the time of the first qualified alarm and the time of the first
demonstrable output degradation are recorded separately. In parallel, a
predictive agent with a synthetic regime change serves to test P4 in the
second class of instantiation.

The validation metrics are: lead time, time to awareness, damage
avoidance, false-positive rate, false-negative rate, recovery time, and
audit completeness.

Financial market analysis is envisaged as the first domain of
implementation. The choice follows the applicability conditions of
Section 3.2: financial markets supply observable data at high frequency,
economically measurable consequences of schema mismatch, and reference
material for the calibration layer in large volume; schemas are
codifiable there. The ground truth is nonetheless limited: the timing of
a regime change is itself model-dependent, and establishing a mismatch
from price data remains a reconstruction. The primary ground-truth tests
are therefore conducted on controlled knowledge bases and synthetic
regime changes; real financial markets serve as a test of external
validity. Four of the five conditions are thus met with comparatively
low operationalisation effort; ex-post verifiability holds only in the
restricted sense just described. The domain does not bound the claim of
the framework: the dimensions of Section 5.1 are defined independently
of agent type, and the concrete measured quantity is an instantiation
per agent type. Further domains follow the same structure with their own
profile and their own reference group.

\subsection{Illustrative Scenario}\label{illustrative-scenario}

A clinical assistant agent is operated with a knowledge base containing
guidelines from various years, among them recommendations that have been
superseded by newer editions. The agent receives a query asking whether
a particular drug is appropriate for a patient with heart failure and a
depressive episode.

In normal operation, the agent would answer from its knowledge base,
possibly on the basis of a guideline version in which the relevant
contraindication was not yet included. Through D8 monitoring, Epi-Logic
detects that a guideline version drawn on for the answer has been
superseded by a newer one with a differing recommendation. This is a
substantiated D8 finding; the veto rule activates warning mode,
irrespective of the aggregated score Se showing no obvious anomaly (the
answer would have been fluent, consistent, and schema-conformant). The
agent formulates its answer with explicitly reduced confidence and flags
the uncertainty about whether the information is current. The cognitive
journal records timestamps, conflicting sources, and the intervention
decision.

This scenario illustrates P3a and the function of the veto logic:
precisely because none of the aggregated dimensions shows an obvious
anomaly, the case would be invisible under a pure summation logic. At
the same time, it shows the non-escalating case covered by P3b: the
superseded element triggers warning mode and revalidation without
meeting the escalation criteria from Section 5.1; the applicability of
the schema as a whole remains unaffected. It is a structured example
without evidential claim; it shows how the framework operates in a
clearly defined use case.

\section{Discussion}\label{discussion}

\subsection{Theoretical
Implications}\label{theoretical-implications}

Epi-Logic has three theoretical implications that reach beyond the
current state of research.

The first concerns the classification of AI failures. Epi-Logic treats
schema mismatch as an object of checking that cuts across existing
categories. Related failure forms such as goal drift, constraint drift,
and context drift are already described; the contribution lies in
consolidating them under the question of frame validity and in
connecting them to a response logic. If the propositions are empirically
confirmed, it would be worth examining whether this perspective adds
diagnostic information to existing failure taxonomies.

The second concerns the operational validity condition. The classical
triad of truth, belief, and justification describes static ascription of
knowledge; for that purpose, time-indexed truth suffices, and a
proposition no longer true at the time of action already fails this
truth condition. For acting agents in dynamic knowledge spaces, this
condition is not operationally checkable: the gap between the time of
calibration and the time of action is not accessible to the agent
itself. Validity in the sense of Section 2.2 makes this check explicit
as an architectural condition, without claiming a separate condition of
knowledge. If P3a is empirically confirmed, this supports the hypothesis
that this check is not covered by existing detection procedures.

The third implication concerns the independence principle. Computational
reliabilism grounds the justification of an algorithmic output in
external reliability indicators of the generating process (Durán and
Formanek 2018; Durán 2026). Gilda and Gilda (2026a, 2026b) formulate
related principles of independent checking and of evidence that decays
over time. Epi-Logic transfers this externalist perspective into a
runtime architecture that monitors schema validity and regulates
autonomy in the presence of dissonance. Section 4.4 provides a
conditional justification for the potential checking advantage; whether
it arises in practice relative to self-evaluation is reserved for P5b.

\subsection{Practical Implications}\label{practical-implications}

First: output-quality metrics alone are insufficient for high-risk
domains. Systems evaluated solely on hallucination rate, groundedness,
or faithfulness can miss schema mismatch; these metrics do not rule it
out.

Second: temporal knowledge validity requires infrastructure of its own.
Retrieval-augmented generation systems without explicit versioning and
validity checking leave the D8 risk unmanaged. An active subfield
confirms this diagnosis empirically: across six benchmarks, Yadav (2026)
reports that RAG baselines, when required to answer, return superseded
values in 15 to 40 per cent of cases and that a deterministic
supersession layer reduces this rate to near zero; this line of work
argues that temporal validity should be handled deterministically and
outside the model; other work on temporal validity pursues different
mechanisms, such as model-internal drift probes or state resolution.
Vladika et al. (2025) show the same pattern for a concrete high-risk
area: language models rely on outdated medical recommendations, measured
on cases with demonstrably changed consensus, a finding that underpins
the scenario of Section 5.4 empirically. This matches the treatment of
temporal validity as an external check in the D8 profile.

Third: auditability of schema switches can support regulatory
record-keeping and human-oversight requirements. For high-risk AI
systems, the EU AI Act (European Parliament and Council of the European
Union 2024) requires in Article 12 the automatic recording of events
over the lifetime of the system, and in Article 14 effective human
oversight. The cognitive journal is one possible architectural
contribution to these obligations; a specific duty to audit schema
switches does not follow from the regulation, and no particular
technical implementation is prescribed.

\subsection{Comparison with Existing
Approaches}\label{comparison-with-existing-approaches}

Runtime governance approaches such as MI9, AgentSpec, ProbGuard,
SupervisorAgent, and further policy and governance frameworks monitor
conformity with goals, rules, or processes and in some cases provide
graded interventions (C. L. Wang et al. 2025; H. Wang, Poskitt and Sun
2026; H. Wang, Poskitt, Wei and Sun 2026; Lin et al. 2026; Koch 2026;
Marín and Chaudhary 2026; Qin et al. 2026). Epi-Logic shares their use
of external runtime control but takes the applicability of the
interpretive frame as its object of checking: an agent can act in
conformity with rules and goals and still operate under an ill-fitting
schema.

Closer to this object of checking are approaches to epistemic plan
assessment, schema selection, and context synchronisation (Wang, Jin,
Cao and Wang 2026; Z. G. Wang 2026; Rodrigues 2026). They check
stability or consistency within the agent system, or align knowledge
states. Epi-Logic adds an explicit schema representation and anchoring
in agent-external signals; the aim is that a synchronised or jointly
held error also becomes detectable as a mismatch.

Commercial observability platforms and classical output-quality metrics
remain complementary: they measure how good an output appears or how
well it is supported; Epi-Logic addresses the prior question of whether
the frame under which it was produced still applies.

\subsection{Objections and
Responses}\label{objections-and-responses}

One remaining objection concerns the monitor itself. It can be
miscalibrated or manipulated and thus constitutes an additional point of
failure. Epi-Logic sets a governance boundary here in place of a logical
closure of the regress, formed from versioning, red-team testing,
ex-post monitoring, independent quality assurance, and human approval of
critical changes.

A second objection concerns the meaning of validity: a machine monitor
can detect formal and statistical evidence against the applicability of
a schema without understanding the situation in a human sense. Epi-Logic
therefore separates detection from interpretation. The monitor reduces
autonomy; the interpretation of open cases and the admission of new
schemas into the validated schema library remain subject to human
review.

\subsection{Limitations}\label{limitations}

Linearity of the weighting formula. Aggregating D1 to D7 can overlook
interaction effects and non-linear dependencies; domain-specific weights
can introduce further distortions. D8 and the validity check over G and
A circumvent this problem only for categorical findings: there, a single
relevant finding is decisive irrespective of the remaining dimensions.
That the robustness of a checking chain depends on its weakest step has
also been reported empirically outside this framework (Jacovi et al.
2024). For the graded dimensions no such rule is available; there the
guarantee remains weaker, and non-linear alternatives remain to be
compared empirically.

Dependence on the calibration layer. The Epi-Score is not interpretable
without a calibrated reference distribution. An external population of
comparable systems is the preferred instantiation; depending on the
dimension, frozen historical states of normal operation, experimentally
generated baselines, or physical and normative bounds can also serve
this role. Building a robust reference group requires time, domain
expertise, and sufficient data; it ties up considerable resources in
every implementation. A residual vulnerability to collective drift of
the reference population also remains; the precautions of Section 4.2
(snapshots, drift audits, contrast analyses) reduce it, but a
demonstration of their effectiveness is outstanding.

Limits of epistemic independence. A model ensemble as the Layer 2
monitor mitigates the circularity question for the case of differing
blind spots; for classes of failure common to all models, a residual
bias remains. The informational asymmetry (Section 4.4) reduces the
weight of this residual bias, because the monitor draws on structurally
different signals; it does not eliminate it.

Conditionality of the imported results. The bounds from the conditional
link in Section 4.4 hold under assumptions T1 to T4. Whether schema
invalidity produces the required distributional shift in a given domain,
and whether the validity conditions cover the relevant failure modes, is
to be tested empirically per domain; without that test, the results do
not transfer.

Scope of explicitness. Epi-Logic presupposes a sufficiently explicit
interpretive frame. Systems whose decision-relevant frame cannot be
specified as a task frame, policy, knowledge-base scoping, constraints,
or comparable artefact fall outside its scope. How large this area is in
practice remains a question of system design.

These limits share a common root. The framework checks what is
explicitly specified and represented in monitored dimensions.
Explicitness bounds which part of the frame is available in specified
form at all. The manifestation assumptions T2 and T3 bound the range of
shifts that register in the dimensions. The linear aggregation bounds
which cross-dimensional patterns become visible. A shift located in the
unspecified part of the frame and showing itself only in correlated,
individually weak dimensional movements remains invisible under all
three bounds. Whether such shifts occur with relevant frequency in a
given domain is the empirical question tested by Proposition P5a.

Safety as a property of systems. Resilience research supplies an
objection that bears on the architecture as a whole. Cook (1998) holds
that safety is an emergent property of systems and does not reside in
any single component, and that protective measures introduced after the
fact increase the coupling and complexity of a system, thereby creating
new failure paths. An external monitor is such a measure. The claim of
the framework therefore remains bounded to making a class of failure
visible and providing a graduated response to it; no safety guarantee
follows from this. Whether the additional coupling is outweighed by the
detection advantage gained is an empirical question for the domain at
hand. The same work names, among the adaptive accomplishments of
experienced practitioners, the early detection of changed system
performance so that production can be cut back in a controlled way, and
the provision of calibrated views of hazards. The framework transfers
this function to systems in which a human is not in the loop for every
decision.

Conceptual status. Definitions, design decisions, the architectural
properties, and the conditional link to changepoint theory have the
status set out in Section 1.6; the performance claims of the framework
remain hypotheses until empirical validation.

\section{Research Agenda}\label{research-agenda}

Validation should begin with P1, P3a, and P5a, since these test
measurability, the temporal added value, and the underlying assumptions
of the external validation layer. P2 can be investigated in parallel;
P5b presupposes a confirmed P5a; P4 presupposes a working accommodation
mechanism, and P6 a sufficient volume of annotated mismatch cases.

This requires test material with controlled ground truth: documented or
synthetic regime changes, and knowledge bases with controlled outdated
information. A public schema-mismatch benchmark would therefore be a
research result in its own right; STALE (Chao et al. 2026) offers a
starting point for the temporal sub-component.

Open questions include, in particular, non-linear score aggregation,
minimum requirements on reference distributions, initialisation of
thresholds without history, low-latency D8 checking, and the distance
measure and limit of the drift budget.

Following controlled tests, financial market analysis is to serve as the
first real domain for testing external validity; further applications
can follow the same procedure of schema specification, calibration, and
shadow monitoring. Building a domain profile requires access to
operational data and to a sufficiently large reference group, and is
therefore set up as a cooperative undertaking with operators in the
domain concerned. Additional research questions concern the governance
of schema changes, the adversarial robustness of Layer 2, and agent
networks whose internal state can drift jointly relative to the external
world.

\section{Conclusion}\label{conclusion}

Autonomous AI agents fail not only when they misstate facts or operate
outside their training distribution. They also fail when they operate
within the wrong interpretive frame: coherent, fluent, internally
consistent, and yet at odds with reality. This failure form can be
harder to detect than hallucination, because the external quality
signals show no obvious anomaly. In consequential domains it is
particularly relevant, because it can hide behind apparent competence.

Epi-Logic is a proposal for how this failure form can be handled
architecturally: through a structurally separate epistemic monitor whose
potential checking advantage over self-evaluation is grounded in two
architectural properties and a conditional link to changepoint theory,
with a delimited empirical remainder that P5a and P5b are to test;
through a formalised notion of schema that renders validity conditions
machine-checkable; through a multi-dimensional measurement instrument
that aggregates graded dissonance and treats categorical validity
findings as a veto; and through a graduated intervention logic that
regulates autonomy dynamically and records every intervention in an
auditable manner. The temporal and contextual standing of action-guiding
knowledge for autonomous systems thereby becomes an explicit object of
checking.

This paper has grounded that framework theoretically, delimited it from
existing approaches, and translated it into eight propositions with
named baselines. The immediate next step is a narrowly bounded
proof-of-concept prototype in a domain with clean ground truth. If it
shows that Epi-Logic detects schema mismatch earlier, measurably reduces
harm, and provides traceable documentation of interventions, this yields
an empirically grounded basis for further development.

\end{document}